\documentclass[10pt]{article}
\usepackage{preprint}

\definecolor{ourrow}{HTML}{EAF1FA}

\newcommand{\ours}[1]{\textbf{#1}}
\newcommand{\method}{SpotAttention}
\newcommand{\selector}{selector}
\newcommand{\Selector}{Selector}
\newcommand{\KLfull}{\mathcal{L}_{\mathrm{KL}}^{\mathrm{Dense}}}
\newcommand{\KLsel}{\mathcal{L}_{\mathrm{KL}}^{\mathrm{Sparse}}}
\newcommand{\Ktrain}{K_{\mathrm{train}}}

\graphicspath{{../}{./}}

\title{SpotAttention: Plug-In Block-Sparse Routing for Pretrained Long-Context Transformers}
\author{%
  \textbf{Huzama Ahmad}\correspondence{huzama@huzama.com} \quad \textbf{Se-Young Yun} \\[3pt]
  \normalsize KAIST%
}

\begin{document}
\maketitle

\begin{abstract}
Long contexts have become standard in pretrained LLMs, yet they remain expensive to run: prefill compute grows quadratically with sequence length, and every decode step re-reads a key-value cache that grows linearly with it. Sparse attention cuts these costs by attending only to a relevant subset of past tokens, but selecting that subset is itself expensive. We present \method{}, a lightweight \selector{} that attaches to a frozen pretrained transformer and learns by KL distillation to estimate its attention distribution. The \selector{} picks the top-$K$ keys each query attends to, and because its estimate is a calibrated distribution, a dual top-$p$ rule reads the per-query, per-layer budget directly from it. Across Qwen3 (dense, 4B–32B) and Qwen3.5 (hybrid linear/full attention, 4B–9B), \method{} matches dense accuracy at contexts up to 128K tokens, eight times the training length. Decode at $L=128$K runs $3.9\times$ faster than FlashAttention and $1.8\times$ faster than Twilight, the strongest training-free baseline. Quantizing the \selector{}'s K-cache to INT4 or FP4 microscale shrinks it $3.5\times$ at no accuracy cost.
\end{abstract}

\section{Introduction}
\label{sec:intro}

Long context has become one of the defining capabilities of modern large language models, letting them follow long documents, work across entire codebases, and ground their answers in large amounts of evidence. The same trend holds for multimodal models, where images and especially video expand into long token sequences. In just a few years, context windows have grown from a few thousand tokens to more than a million \citep{geminiteam2024gemini15}, and they continue to grow.

Self-attention makes long context expensive \citep{vaswani2017attention}. At prefill, its compute grows quadratically with sequence length; at decode, every new token re-reads the entire KV cache, whose size grows linearly with the context and soon dominates memory. FlashAttention \citep{dao2022flashattention} made attention far more IO-efficient and removed the need to materialize the full attention matrix in memory, but it changed neither the quadratic prefill nor the growing per-step read. Attention therefore dominates both latency and memory at long context, and serving these models becomes slow and costly.

\begin{figure}[h]
\centering
\includegraphics[width=0.6\linewidth]{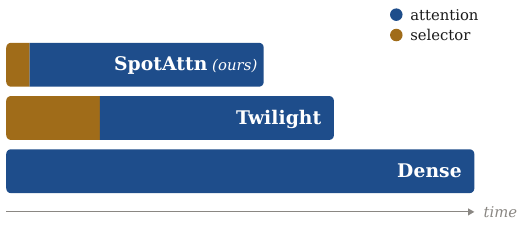}
\caption{\textbf{Cost of sparse attention.} Per-step decoding cost decomposed into \emph{selector} and \emph{attention}. Dense pays for full attention with no selector. Twilight cuts the attention bar but pays a heavy selector. \method{} keeps the same lighter attention with a $3\times$ cheaper learned \selector{}.}
\label{fig:teaser}
\end{figure}

Recently, a growing body of work has shown that attention in trained LLMs is highly redundant: each query attends to only a small fraction of past tokens \citep{liu2023dejavu, xiao2024streamingllm}. Sparse-attention methods exploit that redundancy by attending to a selected subset of tokens, cutting the per-step read from the KV cache. \citet{tang2024quest}, for example, scores 16-token pages with min/max key bounds, and \citet{lin2025twilight} adds top-$p$ pruning on top of that selector. These methods do reduce the attention cost, but selecting which tokens to attend to is itself expensive, and the selector becomes the new bottleneck. DeepSeek Sparse Attention (DSA) \citep{deepseek2025v32} answers it with the \emph{lightning indexer}: a small module trained jointly with the backbone that estimates each layer's attention distribution, so the layer attends only to its top-$K$ keys.

In this paper, we propose \method{}, a learned \selector{} that retrofits onto an already-trained backbone. We adapt the lightning-indexer architecture from DSA: the \selector{} is trained alone via a KL loss against the dense attention, and scores blocks of queries and keys rather than individual tokens, so selection runs as a single tensor-core matmul. We introduce a dual top-$p$ rule that reads the per-query, per-layer budget directly from the \selector{}'s estimated attention distribution, with sink and recency blocks reserved so they do not dominate the nucleus. Twilight \citep{lin2025twilight} also adapts the budget per query, but as a separate top-$p$ stage layered onto Quest's selector, paying the selector cost twice, while \method{}'s nucleus is read from the distribution the \selector{} already produces. Figure~\ref{fig:teaser} previews the cost gap. At $L = 128$K decode on Qwen3-8B, \method{} reaches $3.9\times$ the throughput of FlashAttention and $1.8\times$ that of Twilight (Section~\ref{sec:results-latency}); INT4 or FP4 microscale quantization on the \selector{}'s K-cache shrinks it $3.5\times$ while maintaining BF16 accuracy. We validate \method{} on five backbones across two architecturally distinct families, Qwen3 \citep{yang2025qwen3} (dense attention) and Qwen3.5 (hybrid linear/full attention) at 4B--32B parameters, and show that it matches dense accuracy at contexts up to 128K (Section~\ref{sec:results}).

\paragraph{Contributions.}
\begin{itemize}
  \item \textbf{Plug-in block-sparse attention for pretrained transformers.} A small \selector{} attached to every full-attention layer, trained alone by KL distillation against the dense attention, runs as a single tensor-core matmul.
  \item \textbf{Dual top-$p$ on the learned selection distribution.} We introduce a dual top-$p$ rule that reads the per-query, per-layer budget directly from the \selector{}'s estimated attention distribution, with sink and recency blocks reserved so they do not dominate the nucleus, and with no separate pruning stage. It matches static top-$K$ accuracy at a smaller mean budget.
  \item \textbf{\Selector{} training and inference behaviour.} We show that the \selector{} matches each backbone's teacher distribution across all seven backbones we train, from 4B to 32B. Once trained, the \selector{} reveals a large per-layer budget asymmetry that no static $K$ can match, giving the dual top-$p$ rule a mechanism-level justification.
\end{itemize}

\begin{figure}[t]
\centering
\includegraphics[width=\linewidth]{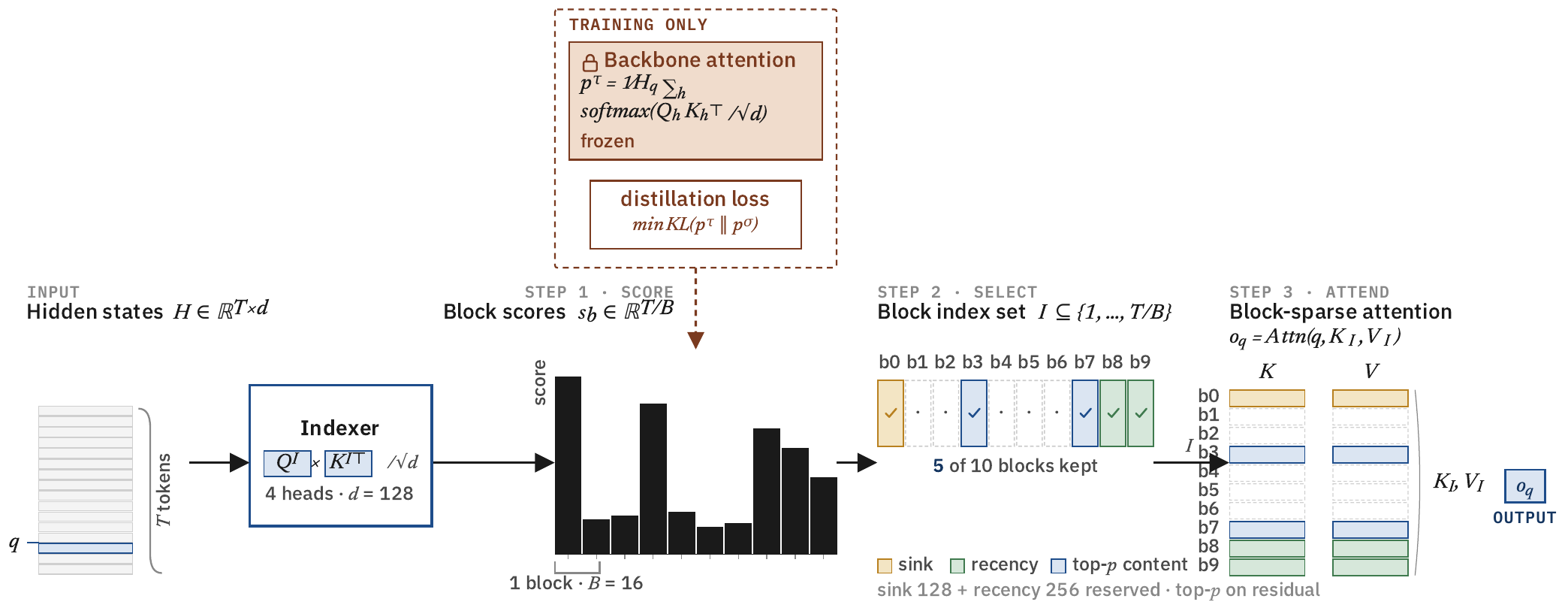}
\caption{\textbf{\method{} for a single query.} The lightweight \emph{Indexer} (a small \selector{}) scores every past block; a dual top-$p$ rule then selects a block index set $I$ comprising a sink prefix, a recency window, and the top-$p$ content blocks, and attention runs only over the gathered $K_I, V_I$. The dashed top band shows the teacher path used during training: the backbone's attention $p^{\tau}$ serves as the target, and the \selector{}'s distribution $p^{\sigma}$ is fit to it by KL divergence.}
\label{fig:method}
\end{figure}

\begin{figure}[t]
\centering
\includegraphics[width=\linewidth]{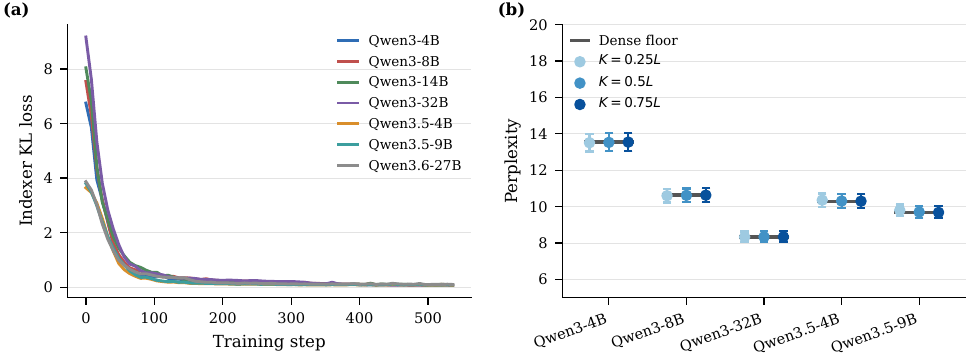}
\caption{\textbf{Training dynamics.} \textbf{(a)}~\Selector{} KL loss vs.\ training step for seven backbones across three families (Qwen3 4B--32B, Qwen3.5-4B/9B, Qwen3.6-27B); every curve falls to near zero within the first ${\sim}200$ steps, so one recipe converges across scale and architecture. \textbf{(b)}~Held-out perplexity for the five evaluation backbones under sparse selection at $K \in \{0.25, 0.5, 0.75\}L$, against the dense floor (gray); sparse perplexity matches the floor at every budget (error bars: $\pm 1$ standard error over held-out 16K blocks).}
\label{fig:training-curves}
\end{figure}

\section{Method}
\label{sec:method}

\method{} attaches a small \selector{} to every full-attention layer of the pretrained backbone, training it by KL divergence to match the layer's head-averaged attention distribution. During inference, the \selector{} scores the past keys of each query and picks a sparse subset, under either a fixed-budget static top-$K$ rule or a content-adaptive dual top-$p$ rule. Figure~\ref{fig:method} illustrates the method for a single query.

\subsection{\Selector{} architecture}
\label{sec:method-selector}

The \selector{} is a tiny multi-head Q-K scorer with $H_{\mathrm{idx}}$ heads of reduced width $d_{\mathrm{idx}}$. For a query at position $q$ and a key at position $t$, it computes a head-weighted ReLU score $s_s[q, t] = \sum_{h=1}^{H_{\mathrm{idx}}} w_{h}\,\mathrm{ReLU}\!\bigl(\mathbf{q}^{I}_{h}\cdot\mathbf{k}^{I}_{h,t}\bigr)$, whose softmax over $t$ gives the estimated attention distribution $p^{\sigma}_{q, t}$. Here $\mathbf{q}^{I}_{h}, \mathbf{k}^{I}_{h,t}$ are the per-head query/key projections and $W_w$ a small linear map producing the per-query mixing weights $w_h$ from the hidden state $\mathbf{h}_q$. Scoring is block-wise: one score per query-block $\times$ key-block tile rather than per token, so selection runs as a single tensor-core matmul.

\subsection{Training objectives}
\label{sec:method-training}

The \selector{} is trained by KL divergence to match the backbone's attention, the teacher path shown in the dashed top band of Figure~\ref{fig:method}. We consider two KL variants that differ in their support: \emph{DenseKL} computes the loss over all $T$ keys, and \emph{SparseKL} computes it only over the \selector{}'s own top-$K$ selected set. Writing $s_t[h, q, \cdot]$ for the backbone's attention scores in head $h$ and $s_s[q, \cdot]$ for the \selector{}'s scores, the teacher $p^{\tau}$ and student $p^{\sigma}$ distributions are

\begin{align}
  p^{\tau}_{b,q,t} &= \tfrac{1}{H_q} \sum_{h=1}^{H_q} \mathrm{softmax}_t\bigl(s_t[h, q, \cdot]\bigr), \\
  p^{\sigma}_{b,q,t} &= \mathrm{softmax}_t\bigl(s_s[q, \cdot]\bigr).
\end{align}

The teacher averages the $H_q$ per-head softmaxes (softmax then mean), matching DSA's construction \citep{deepseek2025v32}; the student is a single softmax over $s_s$.

We train the \selector{} by forward KL from the teacher to the student. \emph{DenseKL} matches the full distribution over all $T$ keys:

\begin{equation}
  \KLfull = \frac{1}{BL} \sum_{b,q} \sum_{t=1}^{T}
    p^{\tau}_{b,q,t} \log \frac{p^{\tau}_{b,q,t}}{p^{\sigma}_{b,q,t}}.
\label{eq:kl-full}
\end{equation}

\emph{SparseKL} matches only over the \selector{}'s own top-$K$ selected set $\mathcal{S}_q$, with both distributions renormalized to that set:

\begin{equation}
  \KLsel = \frac{1}{BL} \sum_{b,q} \sum_{t \in \mathcal{S}_q}
    \widetilde{p}^{\tau}_{b,q,t} \log \frac{\widetilde{p}^{\tau}_{b,q,t}}{\widetilde{p}^{\sigma}_{b,q,t}},
\label{eq:kl-sel}
\end{equation}

where $\widetilde{p}$ denotes a distribution restricted and renormalized to $\mathcal{S}_q$. The KL variant chosen at training time determines the shape of the learned distribution, which controls how far dynamic selection extrapolates (Section~\ref{sec:analysis-distribution}).

The total training loss is the KL alone:
\begin{equation}
  \mathcal{L} \;=\; \mathcal{L}_{\mathrm{KL}}, \qquad \mathcal{L}_{\mathrm{KL}} \in \{\KLfull,\,\KLsel\}.
\label{eq:loss-total}
\end{equation}

\subsection{Selection modes}
\label{sec:method-inference}

At inference, the \selector{} scores every past block for each query, and the layer attends only to keys in a selected subset $\mathcal{S}_q \subseteq \mathcal{B}$ of blocks. Both modes use the per-block maximum $\bar{s}_s[q, b] = \max_{t \in b} s_s[q, t]$, and differ only in how they pick $\mathcal{S}_q$ from those values.

\paragraph{Static top-$K$.} A fixed budget of $K$ blocks per query,
\begin{equation}
  \mathcal{S}_q \;=\; \mathop{\mathrm{argTop}\text{-}K}_{b \in \mathcal{B}}\, \bar{s}_s[q, b].
\label{eq:select-topk}
\end{equation}
The fixed budget measures how well the \selector{} ranks blocks against the teacher it learned from, and serves as the baseline for the adaptive mode below. A single $K$ cannot match the per-layer budget asymmetry the \selector{} reveals (Section~\ref{sec:analysis-perlayer}), which motivates dual top-$p$.

\paragraph{Dual top-$p$ (dynamic budget).}
Dual top-$p$ sets the budget from the \selector{}'s own distribution, in two tiers. First, a fixed prefix of \emph{sink} blocks $\mathcal{S}_{\mathrm{sink}}$ and a fixed suffix of \emph{recency} blocks $\mathcal{S}_{\mathrm{rec}}$ \citep{xiao2024streamingllm} are reserved: these carry a large share of attention mass at any length. Second, nucleus selection \citep{holtzman2019nucleus} runs on the residual distribution
\begin{equation}
  \widetilde{p}^{\sigma}_{q, b} \;=\; \mathop{\mathrm{softmax}}_{b \in \mathcal{B}_{\mathrm{res}}} \bar{s}_s[q, b],
\label{eq:select-nucleus}
\end{equation}
with $\mathcal{B}_{\mathrm{res}} = \mathcal{B} \setminus (\mathcal{S}_{\mathrm{sink}} \cup \mathcal{S}_{\mathrm{rec}})$ the residual blocks, yielding the selected set
\begin{equation}
  \mathcal{S}_q \;=\; \mathcal{S}_{\mathrm{sink}} \,\cup\, \mathcal{S}_{\mathrm{rec}} \,\cup\, \mathcal{S}_{\mathrm{nuc}}(q), \qquad |\mathcal{S}_q| \ge K_{\min},
\label{eq:select-topp}
\end{equation}
where $\mathcal{S}_{\mathrm{nuc}}(q)$ is the smallest prefix of $\mathcal{B}_{\mathrm{res}}$, sorted by $\widetilde{p}^{\sigma}_{q, \cdot}$ descending, whose cumulative mass reaches $p$. The floor $K_{\min}$ guards against degenerate-peaky tiles by extending $\mathcal{S}_{\mathrm{nuc}}$ along descending scores when nucleus alone falls short. The budget adapts per query and per layer, with no separate pruning stage.

\section{Experimental setup}
\label{sec:setup}

\paragraph{Backbones and training.}
We train the \selector{} on seven backbones spanning three families: Qwen3 \citep{yang2025qwen3} at 4B/8B/14B/32B, Qwen3.5 at 4B/9B, and Qwen3.6 at 27B. The full long-context accuracy evaluation covers five of these: Qwen3-4B/8B/32B with dense attention and Qwen3.5-4B/9B with hybrid linear/full attention. We use FineWeb-Edu \citep{penedo2024fineweb} for 100M tokens at $L = 16$K context with $\Ktrain = 8$K selected tokens. We optimize with AdamW \citep{loshchilov2019adamw} under a warmup--constant--decay schedule. The full training recipe and all hyperparameters are in Appendix~\ref{app:hparams}.

\paragraph{Evaluation.}
We evaluate accuracy on four long-context benchmarks at context lengths up to 128K: RULER \citep{hsieh2024ruler}, BABILong \citep{kuratov2024babilong} (qa1--qa3), InfiniteBench \citep{zhang2024infinitebench}, and LongBench-v2 \citep{bai2025longbenchv2}; per-task sample counts and length bins are in Appendix~\ref{app:eval-data}. Qwen3 is capped at 32K, its native maximum position; Qwen3.5 runs to the full 128K. We score with exact match after answer extraction. Each model runs in non-thinking mode, with sampling parameters as shown in Appendix~\ref{app:decoding}. We report both selection modes (Section~\ref{sec:method-inference}): static top-$K$ at $K \in \{0.25, 0.5, 0.75\}\,L$ and dual top-$p$ at $p \in \{0.7, 0.8, 0.9\}$. Prefill is dense throughout; sparse selection is applied only at decode.

\paragraph{Baselines.}
We compare \method{} against the dense backbone and two representative training-free baselines: Quest \citep{tang2024quest} and Twilight \citep{lin2025twilight}.

\paragraph{Implementation.}
To isolate the routing algorithm from implementation drift, we implement \method{}, Quest, and Twilight in Triton with their paper-exact configurations, with FlashAttention as the dense baseline \citep{dao2022flashattention}. Latency is measured on a single NVIDIA B200. 

\paragraph{Quantization.}
We also quantize the \selector{}'s queries and key cache to INT8, INT4, FP8, and FP4, and measure the resulting accuracy and memory/latency trade-offs (Section~\ref{sec:analysis-quant}).

\begin{table}[t]
\centering
\footnotesize
\setlength{\tabcolsep}{3pt}
\resizebox{\textwidth}{!}{%
\begin{tabular}{l cc cc cc cc cc cc cc}
\toprule
& \multicolumn{2}{c}{$L = 8$K} & \multicolumn{2}{c}{$L = 16$K} & \multicolumn{2}{c}{$L = 32$K} & \multicolumn{2}{c}{$L = 64$K} & \multicolumn{2}{c}{$L = 128$K} & \multicolumn{2}{c}{$L = 256$K} & \multicolumn{2}{c}{$L = 512$K} \\
\cmidrule(lr){2-3}\cmidrule(lr){4-5}\cmidrule(lr){6-7}\cmidrule(lr){8-9}\cmidrule(lr){10-11}\cmidrule(lr){12-13}\cmidrule(lr){14-15}
Method & Sel. & Attn & Sel. & Attn & Sel. & Attn & Sel. & Attn & Sel. & Attn & Sel. & Attn & Sel. & Attn \\
\midrule
FlashAttention \citep{dao2022flashattention}            & ---  &  0.85 & ---  &  1.10 & ---  &  1.61 & ---  &  2.62 & ---  &  4.69 & ---  &  8.74 & \multicolumn{2}{c}{OOM} \\
Twilight \citep{lin2025twilight}                        & 0.37 &  0.22 & 0.66 &  0.32 & 1.22 &  0.42 & 2.35 &  0.84 & 4.59 &  1.49 & \multicolumn{2}{c}{OOM} & \multicolumn{2}{c}{OOM} \\
\midrule
\rowcolor{ourrow} \ours{\method{} (top-$K{=}0.5L$)}     & \textbf{0.24} &  0.29 & \textbf{0.29} &  0.47 & \textbf{0.53} &  0.83 & 1.12 &  2.28 & \textbf{1.55} &  2.97 & \textbf{2.87} &  5.81 & \textbf{5.52} & 11.41 \\
\rowcolor{ourrow} \ours{\method{} (top-$p{=}0.7$)}      & \textbf{0.24} &  \textbf{0.18} & \textbf{0.29} &  \textbf{0.21} & \textbf{0.53} &  \textbf{0.29} & \textbf{0.84} &  \textbf{0.76} & \textbf{1.55} &  \textbf{1.32} & 2.88 &  \textbf{2.04} & 5.54 &  \textbf{4.04} \\
\rowcolor{ourrow} \ours{\method{} (top-$p{=}0.9$)}      & \textbf{0.24} &  0.25 & \textbf{0.29} &  0.33 & \textbf{0.53} &  0.56 & \textbf{0.84} &  1.48 & \textbf{1.55} &  2.83 & \textbf{2.87} &  4.63 & 5.53 &  9.18 \\
\midrule
\rowcolor{ourrow} \method{} ($K{=}0.5L$, INT4)          & 0.32 &  0.29 & 0.41 &  0.47 & 0.60 &  0.83 & 1.04 &  1.54 & 1.91 &  2.93 & 3.65 &  5.75 & 7.06 & 11.28 \\
\rowcolor{ourrow} \method{} ($K{=}0.5L$, FP4)           & 0.31 &  0.29 & 0.46 &  0.47 & 0.72 &  0.83 & 1.32 &  1.55 & 2.51 &  2.96 & 4.87 &  5.80 & 9.53 & 11.37 \\
\bottomrule
\end{tabular}
}
\caption{\textbf{Per-token decode latency on Qwen3-8B (ms).} \emph{Sel.} = time the \selector{} takes to decide which keys to attend to; FlashAttention has no \selector{} step. \emph{Attn} = time to run attention on the selected (or full) KV cache. \emph{Attn} is essentially unchanged across precisions because quantization affects only the \selector{}'s K-cache, not the backbone attention. $L > 32$K runs through YaRN factor~16.}
\label{tab:latency-breakdown}
\end{table}

\section{Results}
\label{sec:results}

In this section we trace the \selector{}'s training dynamics and show that its post-training perplexity holds against the dense baseline. We then report the latency breakdown of selector and attention for Qwen3-8B across the $L = 8$K to $L = 512$K sweep, alongside real-world throughput improvements across sequence lengths, and finally accuracy at $K{=}0.5L$ across all five evaluation backbones and generalization to $L = 128$K on Qwen3.5.

\subsection{Training dynamics}
\label{sec:results-training}

Figure~\ref{fig:training-curves}(a) shows the \selector{} trains quickly and the KL loss converging from its initial value to near zero within the first ${\sim}200$ steps, across every backbone we trained. Figure~\ref{fig:training-curves}(b) shows that held-out perplexity under sparse selection sits on the dense floor at every budget we test ($K = 0.25L, 0.5L, 0.75L$), within error bars: sparse selection costs no measurable perplexity. Perplexity is measured over full 16K windows of held-out FineWeb-Edu against the dense floor; because the backbone is frozen and the \selector{} never trains on cross-entropy, the gap from the floor measures the cost of selection alone.

\subsection{Decode latency}
\label{sec:results-latency}

\begin{figure}[t]
\centering
\includegraphics[width=0.6\linewidth]{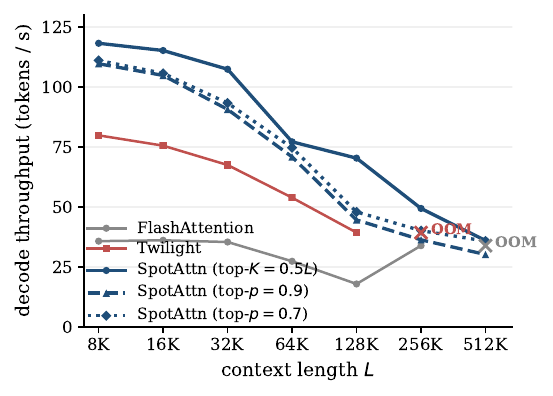}
\caption{\textbf{Decode throughput across context length} on Qwen3-8B. \method{} sustains the highest throughput across all three selection modes from $L = 8$K to $L = 512$K. Twilight OOMs at $L = 256$K, while FlashAttention OOMs at $L = 512$K.}
\label{fig:long-context-decode}
\end{figure}

To isolate the cost of selection, we profile decode latency of the attention kernel and selector kernel on Qwen3-8B against FlashAttention \citep{dao2022flashattention} and Twilight \citep{lin2025twilight} across sequence lengths from $L = 8$K to $L = 512$K, with YaRN factor~16 extending the model beyond its native 32K window. Table~\ref{tab:latency-breakdown} reports the per-step decomposition; Figure~\ref{fig:long-context-decode} reports end-to-end throughput.

\paragraph{Per-step decomposition.}
Table~\ref{tab:latency-breakdown} shows that \method{}'s sparse attention is $1.6$--$3.6\times$ faster than FlashAttention at $L = 128$K depending on selection mode, and the BF16 \selector{} is ${\sim}3\times$ faster than Twilight at the same length, $1.55$~ms vs $4.59$~ms. Quantizing the \selector{}'s K-cache to INT4 or FP4 cuts its memory footprint with no accuracy loss; the per-step \selector{} cost rises slightly ($1.91$~ms INT4 and $2.51$~ms FP4 at $L = 128$K) because the dequantization overhead exceeds the savings from smaller K-cache reads when the BF16 kernel is already memory-light. Twilight runs out of memory at $L = 256$K and FlashAttention at $L = 512$K, where \method{}'s static $K{=}0.5L$ variant still routes in $5.52$~ms.

\begin{table}[t]
\centering
\small
\setlength{\tabcolsep}{6pt}
\begin{tabular}{l c >{\columncolor{ourrow}}c >{\columncolor{ourrow}}c >{\columncolor{ourrow}}c >{\columncolor{ourrow}}c}
\toprule
& & \multicolumn{4}{c}{\ours{\method{}}} \\
\cmidrule(lr){3-6}
Model & Vanilla & $K{=}0.5L$ & $K{=}0.75L$ & top-$p{=}0.7$ & top-$p{=}0.9$ \\
\midrule
Qwen3-4B & 0.543 (0.019) & 0.524 (0.019) & 0.528 (0.018) & 0.505 (0.020) & 0.523 (0.019) \\
Qwen3-8B & 0.537 (0.019) & 0.544 (0.019) & 0.544 (0.019) & 0.525 (0.020) & 0.549 (0.020) \\
Qwen3-32B & 0.599 (0.020) & 0.581 (0.020) & 0.577 (0.019) & 0.516 (0.022) & 0.572 (0.020) \\
Qwen3.5-4B & 0.561 (0.014) & 0.569 (0.014) & 0.557 (0.014) & 0.563 (0.014) & 0.572 (0.014) \\
Qwen3.5-9B & 0.558 (0.015) & 0.566 (0.014) & 0.566 (0.013) & 0.573 (0.013) & 0.566 (0.013) \\
\bottomrule
\end{tabular}
\caption{\textbf{Accuracy parity across backbones.} Each cell is the mean of the per-dataset overall accuracies on RULER, BABILong, and LongBench-v2, with bootstrap standard errors in parentheses (Appendix~\ref{app:bootstrap}); InfiniteBench is excluded so the average is apples-to-apples across Qwen3 (32K cap) and Qwen3.5 (128K). All five backbones land within a single standard error of dense at both static budgets and at dual top-$p{=}0.9$; only the most aggressive setting, top-$p{=}0.7$, trails on the larger Qwen3 backbones.}
\label{tab:accuracy-summary}
\end{table}

\paragraph{End-to-end throughput.}
Figure~\ref{fig:long-context-decode} shows that \method{}'s static $K{=}0.5L$ reaches $107$~tok/s at $L = 32$K ($3.0\times$ FlashAttention, $1.6\times$ Twilight) and $70$~tok/s at $L = 128$K ($3.9\times$ FlashAttention, $1.8\times$ Twilight); at $L = 512$K it sustains $36$~tok/s, where both baselines have run out of memory. Throughput falls only slowly with $L$ because the per-step cost is dominated by the backbone matmul and KV reads. FlashAttention degrades smoothly through $L = 128$K but rebounds at $L = 256$K, which we attribute to a tiled-attention kernel switching at that length. Implementation choices shape end-to-end throughput as much as the algorithm itself, and the same routing scheme can show different speedups across kernels and hardware.

\subsection{Accuracy}
\label{sec:results-accuracy}

At matched budget $K{=}0.5L$, \method{} lands within a single bootstrap standard error of the dense backbone on all five models, and dual top-$p$ at $p{=}0.9$ tracks the static variant; Quest and Twilight at matched budgets on Qwen3-4B sit in the same bootstrap-error band, so the distinguishing axis across selectors is per-step cost, not accuracy. We also show that INT4/FP4 quantization on the \selector{}'s K-cache stays within bootstrap error of BF16 across the Qwen3 family, as shown in Section~\ref{sec:analysis}. Full per-dataset grid, the bootstrap protocol, and quantization results are detailed in Appendices~\ref{app:accuracy-full}, \ref{app:bootstrap}, and~\ref{app:quant}.

\subsection{Training-length invariance}
\label{sec:results-longctx}

\begin{figure}[t]
\centering
\includegraphics[width=\linewidth]{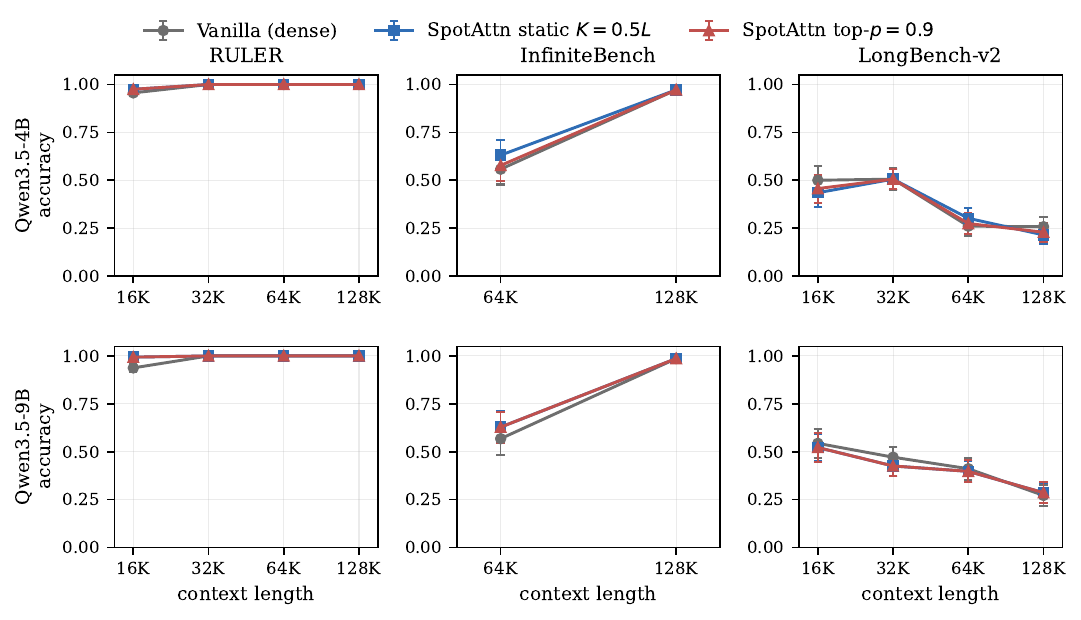}
\caption{\textbf{Training-length generalization.} Mean accuracy across context lengths for \method{} static $K{=}0.5L$ (blue) and \method{} dual top-$p{=}0.9$ (red) follows the dense curve without an accuracy gap.}
\label{fig:longctx}
\end{figure}

We show that a \selector{} trained at fixed sequence length generalizes to any evaluation length. Figure~\ref{fig:longctx} shows accuracy by context length on RULER, InfiniteBench, and LongBench-v2 for the dense backbone, \method{} static $K{=}0.5L$, and \method{} dual top-$p{=}0.9$ on Qwen3.5. Both \method{} variants track the dense curve at every length bin on every dataset, and neither InfiniteBench's 64K--128K span nor LongBench-v2's four bins shows any decay of \method{} relative to dense as length grows.

\section{Analysis and ablations}
\label{sec:analysis}

In this section we analyse the trained \selector{}: what it has learned, what the KL objective shapes it into, and how it tolerates lower-precision storage.

\subsection{Per-layer \texorpdfstring{$K$}{K} asymmetry}
\label{sec:analysis-perlayer}

\begin{figure}[t]
\centering
\includegraphics[width=\linewidth]{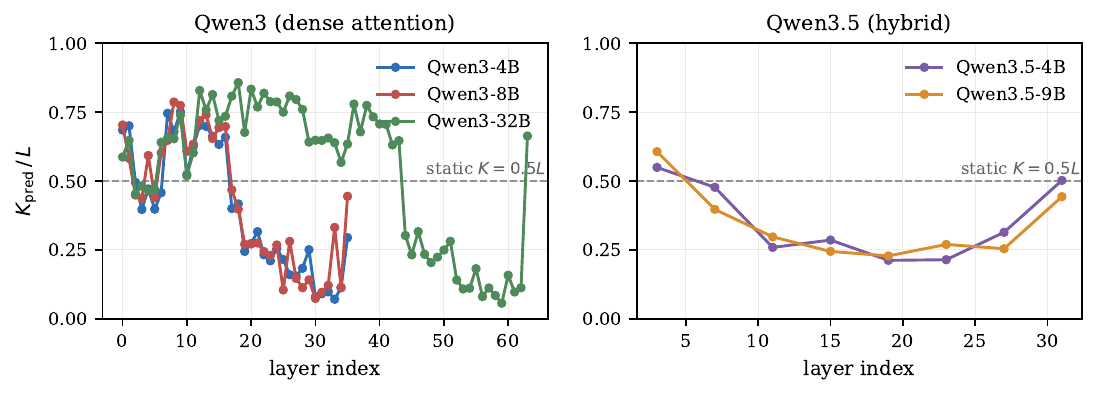}
\caption{\textbf{Per-layer $K$ selected under dual top-$p$.} Mean $K$ per attention layer at $p{=}0.9$ on RULER at 16K context. Dashed line marks the static $K{=}0.5L$ reference. \textbf{Left:} the three dense Qwen3 backbones show early-to-middle layers reaching $K/L \approx 0.7$--$0.85$, while the last third of each network uses very few tokens. \textbf{Right:} Qwen3.5's hybrid backbones carry only eight full-attention layers, and even at that depth the \selector{} picks a U-shaped range.}
\label{fig:perlayer}
\end{figure}

The dual top-$p$ \selector{} chooses how many blocks each layer keeps, and the spread it picks is informative. Figure~\ref{fig:perlayer} plots the mean per-layer $K$ at $p{=}0.9$ on RULER at 16K. In the three dense Qwen3 backbones, early layers ask for $K/L \approx 0.7$--$0.85$ of the context, while layers past two-thirds depth drop to $K/L \approx 0.05$--$0.15$.

The hybrid Qwen3.5 backbones carry only eight full-attention layers, yet those eight still span a U-shape: the asymmetry is not a dense-stack artifact. Appendix~\ref{app:perlayer-grid} shows that the same per-layer signature persists across BABILong, LongBench-v2, and $p \in \{0.7, 0.8, 0.9\}$, so the asymmetry is structural to the backbone rather than a property of the workload or budget.

\subsection{\Selector{} generalization}
\label{sec:analysis-distribution}

\begin{figure}[t]
\centering
\includegraphics[width=0.6\linewidth]{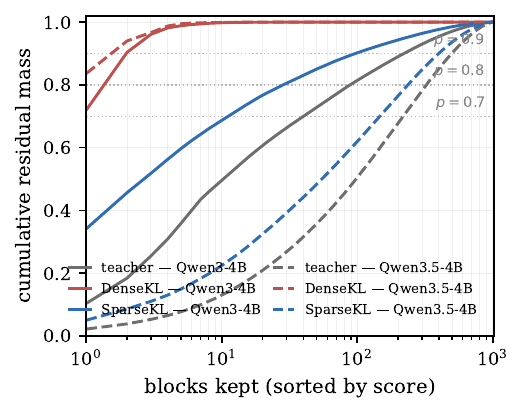}
\caption{\textbf{KL-scope determines \selector{} shape.} Cumulative residual mass against blocks kept in descending score order, measured at the last query. Curves: teacher (gray), DenseKL student (red), SparseKL student (blue). DenseKL hits cumulative mass $0.9$ within ${\sim}10$ blocks; SparseKL tracks the teacher curve.}
\label{fig:distribution}
\end{figure}

\paragraph{Empirical shape gap.}
We trained the \selector{} under DenseKL to test whether it helps the model generalize better or faster than SparseKL. DenseKL turns out razor-sharp: at $L = 32$K on a $L_{\text{train}} = 16$K Qwen3-4B checkpoint, the top-1 block carries a mean of $72\%$ of the residual mass per query, ranging $24\%$--$99\%$ across layers. SparseKL on the same backbone averages $34\%$, ranging $2\%$--$95\%$. On Qwen3.5-4B the gap is wider still: $84\%$ vs $5\%$ on average. The DenseKL cumulative-mass curve in Figure~\ref{fig:distribution} saturates the top-$p$ threshold at $K \approx 10$ blocks regardless of $p$, so peaky distributions cap content reach long before the budget can grow.

\paragraph{Cross-backbone transfer.}
We trained the \selector{} on backbones whose teacher patterns differ in scale and architecture. In every case the SparseKL student tracks the teacher's spread closely; the full heatmap grid is in Appendix~\ref{app:distribution-shape}. One \selector{} geometry therefore produces a distribution that matches whichever backbone it sits on: the recipe transfers shape, not just convergence.

\subsection{Quantized \selector{}: a memory and prefill lever}
\label{sec:analysis-quant}

The \selector{} keeps its own lightweight $K$ cache to compute attention scores. Halving its precision is a constant-factor saving on memory and bandwidth; at decode the backbone KV-cache bandwidth dominates the total, so the per-step \selector{} cost barely moves under quantization. We quantize the \selector{}'s queries and key cache to INT8, INT4, FP8, and FP4 (group-32 microscale) and measure the resulting accuracy and memory/latency trade-offs.

\paragraph{Accuracy.}
Table~\ref{tab:quant} in Appendix~\ref{app:quant} reports Qwen3-4B, -8B, and -32B accuracy under dual top-$p{=}0.9$ with the \selector{}'s queries and key cache quantized. Every variant stays within bootstrap error of the BF16 reference on the dataset-mean column at every backbone. The \selector{} is trained under BF16 and quantized at inference only; the backbone KV cache is unchanged throughout.

\paragraph{Memory and latency.}
The \selector{}'s K-cache scales with $L$, so a precision cut is a linear deployment lever. INT4 / FP4 microscale at group 32 stores $4.5$ bits per element, $0.281\times$ the BF16 footprint: on Qwen3-8B at $L = 128$K the \selector{} K-cache shrinks from $4.50$~GiB at BF16 to $1.27$~GiB at INT4 or $1.20$~GiB at FP4. Per-step \selector{} latency is essentially flat under INT4 at $1.91$~ms versus BF16's $1.55$~ms, and moderately higher under FP4 at $2.51$~ms; the dequantization overhead at decode exceeds the savings from smaller K-cache reads when the BF16 kernel is already memory-light. Single-stream decode throughput follows: at $L = 128$K, INT4 matches BF16 at $70$~tok/s and FP4 drops to $58$~tok/s. The wins from quantization therefore land in memory and in prefill, where the \selector{}'s K-cache size is the dominant cost, rather than in single-stream decode latency.

\section{Related work}
\label{sec:related}

\paragraph{Sparse attention trained with the backbone.}
DeepSeek Sparse Attention \citep{deepseek2025v32}, whose lightning indexer and KL distillation we build on, trains the indexer jointly with the backbone, and the DeepSeek Engram follow-up adds conditional, lookup-based memory as a further axis of sparsity \citep{cheng2026engram}. NSA \citep{yuan2025nsa} and MoBA \citep{lu2025moba} learn block-sparse attention from scratch, while Longformer \citep{beltagy2020longformer} and BigBird \citep{zaheer2020bigbird} fix the pattern entirely, attending over static windows and global tokens. \method{} trains the \selector{} alone, against the dense attention of an already-trained backbone.

\paragraph{Training-free KV-cache selection.}
Many methods select from the KV cache at inference, with no training. Fixed-budget selectors score blocks of keys (Quest \citep{tang2024quest}, InfLLM \citep{xiao2024infllm}), compress the cache during prefill (SnapKV \citep{li2024snapkv}, Ada-KV \citep{feng2025adakv}), pick query-aware patterns (MInference \citep{jiang2024minference}), evict or window tokens (H2O \citep{zhang2023h2o}, StreamingLLM \citep{xiao2024streamingllm}, DuoAttention \citep{xiao2024duoattention}), sample by hashing (MagicPIG \citep{chen2025magicpig}, HashEvict \citep{liu2024hashevict}), or target specific regimes (SpargeAttention \citep{zhang2025spargeattn}, LessIsMore \citep{yang2025lessismore}). Adaptive variants adjust the budget: Twilight \citep{lin2025twilight} layers hierarchical top-$p$ pruning on Quest's selector, Tactic \citep{zhu2025tactic} fits a distribution to an attention-mass target, FlexPrefill \citep{lai2025flexprefill} thresholds per head at prefill, and PyramidKV \citep{cai2024pyramidkv} adapts across layers. These selectors are all hand-designed; \method{}'s is learned, and its distribution is calibrated enough for a nucleus rule to threshold directly. Orthogonal to selection, the cache itself can be compressed via paged management \citep{kwon2023pagedattention} or low-precision storage \citep{hooper2024kvquant,liu2024kivi}, and these compose with \method{}.

\paragraph{Learned retrofit selectors.}
The methods closest to our setting also learn a small selector for a frozen backbone. SeerAttention \citep{gao2024seerattention} trains a pooled attention gate with regression supervision; MISA \citep{misa2026}, concurrent with our work, swaps the lightning indexer for a mixture-of-experts router; both keep a fixed budget. SparQ \citep{ribar2023sparq} approximates the query--key product cheaply, again at a fixed budget and without learning. \method{} distils per-token KL from the dense teacher, which calibrates the \selector{}'s distribution enough to drive a dynamic budget. The Sparse Frontier survey \citep{nawrot2025sparsefrontier} catalogues the representative methods from which we draw our baselines.

\section{Conclusion}
\label{sec:conclusion}

We train a lightweight \selector{} to estimate dense attention distributions and apply a dual top-$p$ rule at inference to pick a variable number of blocks per query and per layer. It learns quickly, transfers across backbones, and matches dense accuracy at every budget we test, generalizing from a 16K training length to 128K evaluation without an accuracy gap. On Qwen3-8B \method{} sustains $70$~tok/s at $128$K, $3.9\times$ faster than FlashAttention and $1.8\times$ faster than Twilight, and reaches $36$~tok/s at $512$K where both baselines have run out of memory. Quantization shrinks the \selector{}'s K-cache $3.5\times$ at no accuracy cost.

\section*{Limitations}
\label{sec:limitations}

Our largest evaluated backbone is 32B and all five evaluation backbones come from the Qwen family on English-only long-context benchmarks; behaviour on other architectures (Llama, Mistral) and non-English data is untested. Sparse-prefill latency is not benchmarked, though the \selector{} trains and runs in the prefill regime, so the path to a sparse-prefill kernel is clear. Dual top-$p$ assumes a well-calibrated \selector{} distribution; under DenseKL training this calibration collapses (Section~\ref{sec:analysis-distribution}), and the rule then under-selects. \Selector{} quantization improves memory and prefill cost but not single-stream decode latency, which is bounded by backbone KV bandwidth; its behaviour under large-batch multi-GPU decode is also not measured.


\bibliography{custom}

\clearpage
\appendix

\section{Hyperparameters}
\label{app:hparams}

Table~\ref{tab:hparams} lists the full \selector{} retrofit recipe. Per-backbone runs vary only the pretrained checkpoint identifier; the \selector{} geometry, optimizer, schedule, and batch arithmetic are identical across all seven backbones.

\begin{table}[h]
\centering
\small
\setlength{\tabcolsep}{5pt}
\begin{tabular}{l l}
\toprule
Hyperparameter & Value \\
\midrule
\multicolumn{2}{l}{\textit{\Selector{} architecture}} \\
heads $H_{\mathrm{idx}}$              & 4 \\
head dim $d_{\mathrm{idx}}$           & 128 \\
RoPE head dim                         & 64 \\
block size $B$                        & 16 \\
score                                 & $\sum_h w_h\,\mathrm{ReLU}(\mathbf{q}^{I}_{h}\cdot\mathbf{k}^{I}_{h})$ \\
selection kernel                      & fused-Triton top-$K$ \\
\midrule
\multicolumn{2}{l}{\textit{Training data}} \\
corpus                                & FineWeb-Edu \texttt{sample-10BT} \\
packing                               & streaming, EOS-delimited \\
sequence length $L$                   & 16{,}384 \\
training tokens                       & $1{\times}10^{8}$ \\
seed                                  & 42 \\
\midrule
\multicolumn{2}{l}{\textit{\Selector{} training}} \\
backbone                              & frozen \\
trainable                             & \selector{} projections only \\
KL objective                          & SparseKL (Eq.~\ref{eq:kl-sel}) \\
$K_{\mathrm{train}}$                  & $8{,}192 = 0.5\,L$ \\
precision                             & bfloat16 \\
\midrule
\multicolumn{2}{l}{\textit{Optimizer \& schedule}} \\
optimizer                             & AdamW \\
$(\beta_1, \beta_2)$                  & $(0.9, 0.95)$ \\
$\epsilon$                            & $10^{-8}$ \\
weight decay                          & $0.1$ \\
peak LR                               & $1{\times}10^{-3}$ \\
min LR                                & $5{\times}10^{-5}$ \\
warmup steps                          & $100$ \\
constant-phase fraction               & $0.30$ \\
decay phase                           & cosine to min LR \\
gradient clip (global $\ell_2$)       & $1.0$ \\
\midrule
\multicolumn{2}{l}{\textit{Batch arithmetic}} \\
per-device batch                      & $2$ sequences \\
gradient accumulation                 & $4$ steps \\
processes (DDP)                       & $1$ GPU \\
effective batch                       & $8$ sequences ($131{,}072$ tokens) \\
optimizer steps                       & $\lceil 10^{8} / 131{,}072 \rceil = 763$ \\
\midrule
\multicolumn{2}{l}{\textit{Inference (selection modes)}} \\
static top-$K$ fractions              & $K \in \{0.25,\, 0.5,\, 0.75\}\,L$ \\
dual top-$p$ masses                   & $p \in \{0.7,\, 0.8,\, 0.9\}$ \\
sink boundary                         & $128$ tokens (absolute prefix) \\
recency window                        & $256$ tokens (per tile) \\
$K_{\min}$ floor                      & $512$ tokens ($32$ blocks) \\
softmax temperature                   & $1.0$ \\
\bottomrule
\end{tabular}
\caption{\textbf{Default hyperparameters for the \selector{} retrofit.} Training settings (top groups) and inference selection settings (bottom group) are identical across all seven backbones; only the pretrained-checkpoint identifier varies per run.}
\label{tab:hparams}
\end{table}

\section{Held-out perplexity protocol}
\label{app:ppl}

We report token-level perplexity, $\mathrm{PPL} = \exp(\overline{\mathrm{CE}})$, with $\overline{\mathrm{CE}}$ the mean next-token cross-entropy.

\paragraph{Data.} We score on FineWeb-Edu \citep{penedo2024fineweb} (the \texttt{sample-10BT} split) through the same streaming, EOS-delimited packing loader as \selector{} training, packed to 16{,}384-token blocks. The only change from training is the shuffle seed: a held-out seed (\texttt{987654321}) distinct from the training seed, so the document ordering differs. Contamination is moot regardless: the backbone is frozen and the \selector{} never optimizes cross-entropy, so no component is fit to this metric.

\paragraph{Scoring.} We score all $L-1$ next-token predictions in each 16K block and forward enough blocks to score at least $65{,}536$ tokens ($\approx 82$K scored tokens per cell at batch size~1).

\paragraph{Forward regime.} Each block is one prefill forward (\texttt{use\_cache=False}), so for the sparse model the \selector{} and block-sparse kernels run exactly as in training. Positions run sequentially $0 \ldots L{-}1$ under a full causal mask, matching training; the loader's per-document position resets are disabled here, since resetting RoPE mid-block while attention still crosses document seams produces incoherent logits.

\paragraph{Operating points.} The sparse series pin the \selector{} budget to $\lfloor cL \rfloor$ rounded to the block size~16, for $c \in \{0.25, 0.5, 0.75\}$: 4{,}096, 8{,}192, and 12{,}288 keys at $L = 16$K. The dense floor runs the same checkpoint with the \selector{} disabled, attending over the full context.

\paragraph{Error bars.} Each 16K block is one independent unit. We take the standard error of the mean cross-entropy as $\mathrm{stdev}/\sqrt{n}$ over blocks and propagate it to perplexity by the delta method, $\mathrm{SE}(\mathrm{PPL}) = \mathrm{PPL} \cdot \mathrm{SE}(\overline{\mathrm{CE}})$.

\section{Evaluation protocol}
\label{app:eval-protocol}

\subsection{Data configuration}
\label{app:eval-data}
Table~\ref{tab:eval-data} summarises the per-benchmark task selection, length bins, and sample counts. RULER and BABILong are evaluated in fixed length bins (16384, 32768, 65536, 131072); InfiniteBench and LongBench-v2 are capped at 131072 tokens. LongBench-v2 uses all available samples. Samples are then capped to each model's maximum context: 32K for Qwen3, 128K for Qwen3.5.

\begin{table}[h]
\centering
\footnotesize
\setlength{\tabcolsep}{4pt}
\begin{tabular}{@{}l p{3.8cm} l c@{}}
\toprule
Benchmark & Tasks & Lengths & Samples \\
\midrule
RULER & all & 16/32/64/128K & 32 \\
BABILong & qa1, qa2, qa3 & 16/32/64/128K & 100 \\
InfiniteBench & passkey, kv\_retrieval, longbook\_choice\_eng, longbook\_qa\_eng & $\le$128K & 64 \\
LongBench-v2 & all & $\le$128K & all \\
\bottomrule
\end{tabular}
\caption{\textbf{Evaluation-data configuration.} ``Samples'' is per task. Lengths 16/32/64/128K $=$ 16384, 32768, 65536, 131072.}
\label{tab:eval-data}
\end{table}

\subsection{Decoding}
\label{app:decoding}
Table~\ref{tab:decoding} lists the per-model sampling settings. All evaluation uses sampling under each model's recommended non-thinking preset; the Qwen3 thinking preset is listed for reference only.

\begin{table}[h]
\centering
\small
\setlength{\tabcolsep}{4pt}
\begin{tabular}{@{}l l cccc@{}}
\toprule
Model & Mode & temp & top-$p$ & top-$k$ & min-$p$ \\
\midrule
Qwen3   & thinking     & 0.6 & 0.95 & 20 & 0.0 \\
Qwen3   & non-thinking & 0.7 & 0.80 & 20 & 0.0 \\
Qwen3.5 & non-thinking & 0.7 & 0.80 & 20 & 0.0 \\
\bottomrule
\end{tabular}
\caption{\textbf{Decoding (sampling) settings.} Evaluation uses the non-thinking presets across all four benchmarks; the Qwen3 thinking preset is listed for reference.}
\label{tab:decoding}
\end{table}

\subsection{Prompts and answer extraction}
\label{app:prompts}

\paragraph{Prompt templates.}
All four benchmarks share the same system prompt and two user templates, dispatched per example as free-text or multiple-choice. Both templates instruct the model to wrap its final answer in \texttt{<answer>...</answer>} so a regex extractor can recover it. The system prompt is
{\footnotesize
\begin{verbatim}
You are a helpful assistant for question-answering
tasks. Always answer based solely on the provided
context.
\end{verbatim}
}
The free-text template (RULER, BABILong, InfiniteBench \texttt{passkey}/\allowbreak\texttt{kv\_retrieval}/\allowbreak\texttt{longbook\_qa\_eng}):
{\footnotesize
\begin{verbatim}
# Context
{context}
# Question
{question}
# Instructions
Answer the question based on the context. Provide
your final answer wrapped in <answer> tags, e.g.
<answer>X</answer>.
\end{verbatim}
}
The multiple-choice template (LongBench-v2, InfiniteBench \texttt{longbook\_choice\_eng}) reuses the context + question blocks and adds a four-option list (\texttt{(A)...(D)}), with the instruction to return the letter inside \texttt{<answer>...</answer>}. Both templates run through \texttt{tokenizer.apply\_chat\_template(..., add\_generation\_prompt=True)} so the chat formatting is the model's own.

\paragraph{Answer extraction.}
We extract the last \texttt{<answer>...</answer>} payload from the completion. For non-thinking models on every benchmark, the opening tag is appended to the prompt so the model resumes generation immediately after it; the same regex still catches the closing tag. In thinking mode, tag-prefilling is suppressed so it does not clobber the model's native \texttt{<think>} block. For content-word answers (BABILong, multiple-choice, free-form F1) we score 0 when no closing tag is found; for unique-token retrieval (RULER NIAH, passkey, kv) we fall back to substring-matching the full completion, since the gold values are random nonces unlikely to match by accident.

\subsection{Scoring}
\label{app:scoring}
Each example is scored by one of three rules:
\begin{itemize}\setlength\itemsep{0pt}
  \item \textbf{Substring} (RULER NIAH, BABILong, InfiniteBench \texttt{passkey} / \texttt{kv\_retrieval}): the score is the fraction of gold answers that appear as case-insensitive substrings of the extracted text. With a single gold this collapses to a 0/1 indicator.
  \item \textbf{SQuAD-style F1} (InfiniteBench \texttt{longbook\_qa\_eng}): token F1 between extracted and gold after lowercasing, stripping articles and punctuation, and whitespace tokenization; maximum across gold candidates.
  \item \textbf{Multiple-choice letter} (LongBench-v2, InfiniteBench \texttt{longbook\_choice\_eng}): the first A--D token in the extracted text, scored 1 if it matches the gold letter and 0 otherwise.
\end{itemize}
Per-example scores enter the bootstrap unchanged.

\subsection{Bootstrap protocol}
\label{app:bootstrap}
Every main-text and appendix accuracy number is a bootstrap mean with bootstrap standard error. We treat each cell, one (model, mode, dataset) triple, as an independent estimation problem. The unit of resampling is the per-example score the evaluator recorded: 0/1 under substring and multiple-choice scorers; $[0, 1]$ under the InfiniteBench QA F1 scorer. For each cell we draw $B = 1000$ resamples with replacement and report the cell mean alongside the standard deviation of the resample means; the seed is fixed so re-runs are bit-stable.

Standard errors of dataset-mean rows (the rightmost columns of every accuracy table) propagate the per-cell standard errors under independence across datasets, which holds because the example sets are disjoint.

Per-cell sample sizes vary by benchmark: RULER $\approx$$160$, BABILong $300$, InfiniteBench $64$--$101$, LongBench-v2 $\approx$$270$ examples per cell. A single standard error spans roughly $\pm 0.015$ at $p{=}0.5$ on RULER, BABILong, and LongBench-v2, and $\pm 0.03$--$0.05$ on InfiniteBench.

\section{Full accuracy results}
\label{app:accuracy-full}

\paragraph{Qwen3 family (4B, 8B, 32B).}
Table~\ref{tab:accuracy-qwen3} reports the per-dataset grid for the Qwen3 family. All sparse selectors land within bootstrap error of the dense backbone at every matched budget. The 32K context cap on this family excludes InfiniteBench, whose bins start at 64K. Quest and Twilight are reported where their runs have completed: Qwen3-4B fully; Qwen3-8B and Qwen3-32B partially.
\begin{table}[t]
\centering
\small
\setlength{\tabcolsep}{6pt}
\begin{tabular}{ll ccc}
\toprule
Model & Method & RULER & BABILong & LongBench-v2 \\
\midrule
\multirow{13}{*}{Qwen3-4B} & Vanilla & 0.981 (0.011) & 0.333 (0.027) & 0.315 (0.049) \\
 & Quest \citep{tang2024quest} @ $0.25L$ & 0.981 (0.011) & 0.317 (0.026) & 0.281 (0.047) \\
 & Quest @ $0.5L$ & 0.981 (0.011) & 0.337 (0.027) & 0.292 (0.048) \\
 & Quest @ $0.75L$ & 0.981 (0.011) & 0.330 (0.028) & 0.281 (0.048) \\
 & Twilight \citep{lin2025twilight} @ $0.25L$ & 0.919 (0.022) & 0.317 (0.026) & 0.281 (0.047) \\
 & Twilight @ $0.5L$ & 0.975 (0.012) & 0.307 (0.027) & 0.281 (0.046) \\
 & Twilight @ $0.75L$ & 0.981 (0.011) & 0.317 (0.026) & 0.270 (0.045) \\
 & \cellcolor{ourrow}\ours{\method{}} static @ $0.25L$ & \cellcolor{ourrow}0.963 (0.015) & \cellcolor{ourrow}0.320 (0.027) & \cellcolor{ourrow}0.270 (0.047) \\
 & \cellcolor{ourrow}\ours{\method{}} static @ $0.5L$ & \cellcolor{ourrow}0.969 (0.014) & \cellcolor{ourrow}0.323 (0.027) & \cellcolor{ourrow}0.281 (0.048) \\
 & \cellcolor{ourrow}\ours{\method{}} static @ $0.75L$ & \cellcolor{ourrow}0.981 (0.011) & \cellcolor{ourrow}0.323 (0.027) & \cellcolor{ourrow}0.281 (0.046) \\
 & \cellcolor{ourrow}\ours{\method{}} top-$p{=}0.7$ & \cellcolor{ourrow}0.931 (0.020) & \cellcolor{ourrow}0.313 (0.027) & \cellcolor{ourrow}0.270 (0.048) \\
 & \cellcolor{ourrow}\ours{\method{}} top-$p{=}0.8$ & \cellcolor{ourrow}0.956 (0.016) & \cellcolor{ourrow}0.330 (0.028) & \cellcolor{ourrow}0.270 (0.047) \\
 & \cellcolor{ourrow}\ours{\method{}} top-$p{=}0.9$ & \cellcolor{ourrow}0.975 (0.013) & \cellcolor{ourrow}0.323 (0.028) & \cellcolor{ourrow}0.270 (0.047) \\
\midrule
\multirow{13}{*}{Qwen3-8B} & Vanilla & 0.988 (0.009) & 0.330 (0.027) & 0.292 (0.048) \\
 & Quest \citep{tang2024quest} @ $0.25L$ & 0.994 (0.006) & 0.333 (0.027) & 0.315 (0.048) \\
 & Quest @ $0.5L$ & 0.988 (0.009) & 0.323 (0.026) & 0.326 (0.051) \\
 & Quest @ $0.75L$ & 0.988 (0.009) & 0.333 (0.028) & 0.326 (0.049) \\
 & Twilight \citep{lin2025twilight} @ $0.25L$ & 0.938 (0.019) & 0.320 (0.027) & 0.315 (0.049) \\
 & Twilight @ $0.5L$ & 0.975 (0.012) & 0.303 (0.026) & 0.326 (0.051) \\
 & Twilight @ $0.75L$ & 0.988 (0.009) & 0.323 (0.028) & 0.326 (0.050) \\
 & \cellcolor{ourrow}\ours{\method{}} static @ $0.25L$ & \cellcolor{ourrow}0.981 (0.011) & \cellcolor{ourrow}0.333 (0.027) & \cellcolor{ourrow}0.315 (0.050) \\
 & \cellcolor{ourrow}\ours{\method{}} static @ $0.5L$ & \cellcolor{ourrow}0.988 (0.009) & \cellcolor{ourrow}0.330 (0.028) & \cellcolor{ourrow}0.315 (0.049) \\
 & \cellcolor{ourrow}\ours{\method{}} static @ $0.75L$ & \cellcolor{ourrow}0.988 (0.009) & \cellcolor{ourrow}0.320 (0.026) & \cellcolor{ourrow}0.326 (0.051) \\
 & \cellcolor{ourrow}\ours{\method{}} top-$p{=}0.7$ & \cellcolor{ourrow}0.931 (0.021) & \cellcolor{ourrow}0.330 (0.028) & \cellcolor{ourrow}0.315 (0.050) \\
 & \cellcolor{ourrow}\ours{\method{}} top-$p{=}0.8$ & \cellcolor{ourrow}0.963 (0.015) & \cellcolor{ourrow}0.343 (0.028) & \cellcolor{ourrow}0.315 (0.049) \\
 & \cellcolor{ourrow}\ours{\method{}} top-$p{=}0.9$ & \cellcolor{ourrow}0.981 (0.011) & \cellcolor{ourrow}0.350 (0.028) & \cellcolor{ourrow}0.315 (0.051) \\
\midrule
\multirow{13}{*}{Qwen3-32B} & Vanilla & 1.000 (0.000) & 0.403 (0.029) & 0.393 (0.054) \\
 & Quest \citep{tang2024quest} @ $0.25L$ & 1.000 (0.000) & 0.390 (0.028) & 0.382 (0.050) \\
 & Quest @ $0.5L$ & 1.000 (0.000) & 0.403 (0.028) & 0.393 (0.053) \\
 & Quest @ $0.75L$ & 1.000 (0.000) & 0.407 (0.028) & 0.382 (0.050) \\
 & Twilight \citep{lin2025twilight} @ $0.25L$ & 0.963 (0.015) & 0.363 (0.028) & 0.371 (0.050) \\
 & Twilight @ $0.5L$ & 0.994 (0.006) & 0.383 (0.028) & 0.404 (0.050) \\
 & Twilight @ $0.75L$ & 1.000 (0.000) & 0.403 (0.028) & 0.393 (0.051) \\
 & \cellcolor{ourrow}\ours{\method{}} static @ $0.25L$ & \cellcolor{ourrow}1.000 (0.000) & \cellcolor{ourrow}0.377 (0.028) & \cellcolor{ourrow}0.348 (0.052) \\
 & \cellcolor{ourrow}\ours{\method{}} static @ $0.5L$ & \cellcolor{ourrow}1.000 (0.000) & \cellcolor{ourrow}0.383 (0.029) & \cellcolor{ourrow}0.360 (0.052) \\
 & \cellcolor{ourrow}\ours{\method{}} static @ $0.75L$ & \cellcolor{ourrow}1.000 (0.000) & \cellcolor{ourrow}0.383 (0.029) & \cellcolor{ourrow}0.348 (0.050) \\
 & \cellcolor{ourrow}\ours{\method{}} top-$p{=}0.7$ & \cellcolor{ourrow}0.838 (0.029) & \cellcolor{ourrow}0.350 (0.027) & \cellcolor{ourrow}0.360 (0.052) \\
 & \cellcolor{ourrow}\ours{\method{}} top-$p{=}0.8$ & \cellcolor{ourrow}0.900 (0.025) & \cellcolor{ourrow}0.353 (0.027) & \cellcolor{ourrow}0.371 (0.050) \\
 & \cellcolor{ourrow}\ours{\method{}} top-$p{=}0.9$ & \cellcolor{ourrow}0.994 (0.006) & \cellcolor{ourrow}0.363 (0.029) & \cellcolor{ourrow}0.360 (0.051) \\
\bottomrule
\end{tabular}
\caption{\textbf{Full accuracy grid: Qwen3 family.} Three Qwen3 backbones (32K context cap) across RULER, BABILong, and LongBench-v2. Static-$K$ rows use $K \in \{0.25, 0.5, 0.75\}\,L$; dual top-$p$ rows use $p \in \{0.7, 0.8, 0.9\}$ and live below the static block on each backbone. Cells are mean accuracy (bootstrap standard error in parentheses, Appendix~\ref{app:bootstrap}) aggregated across length bins the cell ran at. InfiniteBench is excluded because its 64K / 128K bins exceed the Qwen3 context cap.}
\label{tab:accuracy-qwen3}
\end{table}

\paragraph{Qwen3.5 family (4B, 9B).}
Table~\ref{tab:accuracy-qwen35} reports the per-dataset grid for the Qwen3.5 family at the native 128K context. Quest and Twilight have not yet been run on Qwen3.5 and are omitted.
\begin{table}[t]
\centering
\small
\setlength{\tabcolsep}{4pt}
\begin{tabular}{ll cccc}
\toprule
Model & Method & RULER & BABILong & InfiniteBench & LongBench-v2 \\
\midrule
\multirow{7}{*}{Qwen3.5-4B} & Vanilla & 0.964 (0.014) & 0.343 (0.028) & 0.853 (0.033) & 0.377 (0.030) \\
 & \cellcolor{ourrow}\ours{\method{}} static @ $0.25L$ & \cellcolor{ourrow}0.979 (0.011) & \cellcolor{ourrow}0.347 (0.027) & \cellcolor{ourrow}0.872 (0.031) & \cellcolor{ourrow}0.366 (0.030) \\
 & \cellcolor{ourrow}\ours{\method{}} static @ $0.5L$ & \cellcolor{ourrow}0.979 (0.010) & \cellcolor{ourrow}0.363 (0.027) & \cellcolor{ourrow}0.874 (0.030) & \cellcolor{ourrow}0.366 (0.028) \\
 & \cellcolor{ourrow}\ours{\method{}} static @ $0.75L$ & \cellcolor{ourrow}0.979 (0.010) & \cellcolor{ourrow}0.340 (0.028) & \cellcolor{ourrow}0.844 (0.035) & \cellcolor{ourrow}0.351 (0.029) \\
 & \cellcolor{ourrow}\ours{\method{}} top-$p{=}0.7$ & \cellcolor{ourrow}0.979 (0.010) & \cellcolor{ourrow}0.337 (0.027) & \cellcolor{ourrow}0.860 (0.033) & \cellcolor{ourrow}0.373 (0.030) \\
 & \cellcolor{ourrow}\ours{\method{}} top-$p{=}0.8$ & \cellcolor{ourrow}0.979 (0.010) & \cellcolor{ourrow}0.350 (0.028) & \cellcolor{ourrow}0.873 (0.031) & \cellcolor{ourrow}0.380 (0.029) \\
 & \cellcolor{ourrow}\ours{\method{}} top-$p{=}0.9$ & \cellcolor{ourrow}0.979 (0.011) & \cellcolor{ourrow}0.370 (0.029) & \cellcolor{ourrow}0.859 (0.033) & \cellcolor{ourrow}0.366 (0.028) \\
\midrule
\multirow{7}{*}{Qwen3.5-9B} & Vanilla & 0.948 (0.016) & 0.310 (0.027) & 0.866 (0.032) & 0.417 (0.030) \\
 & \cellcolor{ourrow}\ours{\method{}} static @ $0.25L$ & \cellcolor{ourrow}0.995 (0.005) & \cellcolor{ourrow}0.303 (0.027) & \cellcolor{ourrow}0.883 (0.029) & \cellcolor{ourrow}0.395 (0.029) \\
 & \cellcolor{ourrow}\ours{\method{}} static @ $0.5L$ & \cellcolor{ourrow}0.995 (0.005) & \cellcolor{ourrow}0.303 (0.027) & \cellcolor{ourrow}0.883 (0.030) & \cellcolor{ourrow}0.399 (0.031) \\
 & \cellcolor{ourrow}\ours{\method{}} static @ $0.75L$ & \cellcolor{ourrow}0.995 (0.005) & \cellcolor{ourrow}0.303 (0.027) & \cellcolor{ourrow}0.883 (0.030) & \cellcolor{ourrow}0.399 (0.029) \\
 & \cellcolor{ourrow}\ours{\method{}} top-$p{=}0.7$ & \cellcolor{ourrow}0.995 (0.005) & \cellcolor{ourrow}0.337 (0.027) & \cellcolor{ourrow}0.866 (0.033) & \cellcolor{ourrow}0.388 (0.029) \\
 & \cellcolor{ourrow}\ours{\method{}} top-$p{=}0.8$ & \cellcolor{ourrow}0.990 (0.008) & \cellcolor{ourrow}0.357 (0.028) & \cellcolor{ourrow}0.894 (0.028) & \cellcolor{ourrow}0.391 (0.030) \\
 & \cellcolor{ourrow}\ours{\method{}} top-$p{=}0.9$ & \cellcolor{ourrow}0.995 (0.006) & \cellcolor{ourrow}0.303 (0.026) & \cellcolor{ourrow}0.883 (0.029) & \cellcolor{ourrow}0.399 (0.029) \\
\bottomrule
\end{tabular}
\caption{\textbf{Full accuracy grid: Qwen3.5 family.} Two hybrid Qwen3.5 backbones (128K native context) across the four long-context benchmarks. Static-$K$ rows use $K \in \{0.25, 0.5, 0.75\}\,L$; dual top-$p$ rows use $p \in \{0.7, 0.8, 0.9\}$. Cells are mean accuracy (bootstrap standard error in parentheses, Appendix~\ref{app:bootstrap}) aggregated across length bins the cell ran at. Quest and Twilight have not yet been run on this family; the Qwen3 family table is the only complete head-to-head against training-free baselines.}
\label{tab:accuracy-qwen35}
\end{table}

\paragraph{Per-subtask resolution.}
Table~\ref{tab:per-task} reports subtask-level accuracy on Qwen3-4B against \method{} static $K{=}0.5L$ and dual top-$p{=}0.9$. The family-table dataset rows above are example-weighted means over these subtasks.
\begin{table}[t]
\centering
\footnotesize
\setlength{\tabcolsep}{4pt}
\renewcommand{\arraystretch}{1.0}
\begin{tabular}{l c >{\columncolor{ourrow}}c >{\columncolor{ourrow}}c}
\toprule
Task & Vanilla & \ours{$K{=}0.5L$} & \ours{$p{=}0.9$} \\
\midrule
\multicolumn{4}{l}{\textit{RULER}} \\
niah\_multikey\_1 & 1.000 (0.000) & 1.000 (0.000) & 1.000 (0.000) \\
niah\_multikey\_2 & 1.000 (0.000) & 1.000 (0.000) & 1.000 (0.000) \\
niah\_multikey\_3 & 0.957 (0.043) & 0.870 (0.070) & 0.913 (0.060) \\
niah\_single\_1 & 1.000 (0.000) & 1.000 (0.000) & 1.000 (0.000) \\
niah\_single\_2 & 1.000 (0.000) & 1.000 (0.000) & 1.000 (0.000) \\
niah\_single\_3 & 0.939 (0.041) & 0.939 (0.042) & 0.939 (0.040) \\
\midrule
\multicolumn{4}{l}{\textit{BABILong}} \\
babilong\_qa1 & 0.680 (0.047) & 0.710 (0.045) & 0.700 (0.046) \\
babilong\_qa2 & 0.170 (0.037) & 0.180 (0.040) & 0.180 (0.039) \\
babilong\_qa3 & 0.150 (0.037) & 0.080 (0.028) & 0.090 (0.029) \\
\midrule
\multicolumn{4}{l}{\textit{LongBench-v2}} \\
Code Repository & 0.000 (0.000) & 0.000 (0.000) & 0.000 (0.000) \\
Long ICL & 0.200 (0.173) & 0.200 (0.176) & 0.200 (0.181) \\
Long Structured Data & 0.000 (0.000) & 0.000 (0.000) & 0.000 (0.000) \\
Long-dialogue Hist. & 0.250 (0.122) & 0.167 (0.106) & 0.083 (0.076) \\
Multi-Doc QA & 0.368 (0.113) & 0.316 (0.107) & 0.316 (0.107) \\
Single-Doc QA & 0.347 (0.065) & 0.327 (0.068) & 0.327 (0.069) \\
\bottomrule
\end{tabular}
\caption{\textbf{Per-task accuracy on Qwen3-4B.} Vanilla against \method{} static $K{=}0.5L$ and dual top-$p{=}0.9$ on every subtask of RULER, BABILong, and LongBench-v2; mean over all length bins the model ran at, with bootstrap standard error in parentheses (Appendix~\ref{app:bootstrap}).}
\label{tab:per-task}
\end{table}

\section{Per-layer \texorpdfstring{$K$}{K} across datasets and budgets}
\label{app:perlayer-grid}

\begin{figure}[t]
\centering
\includegraphics[width=0.9\linewidth]{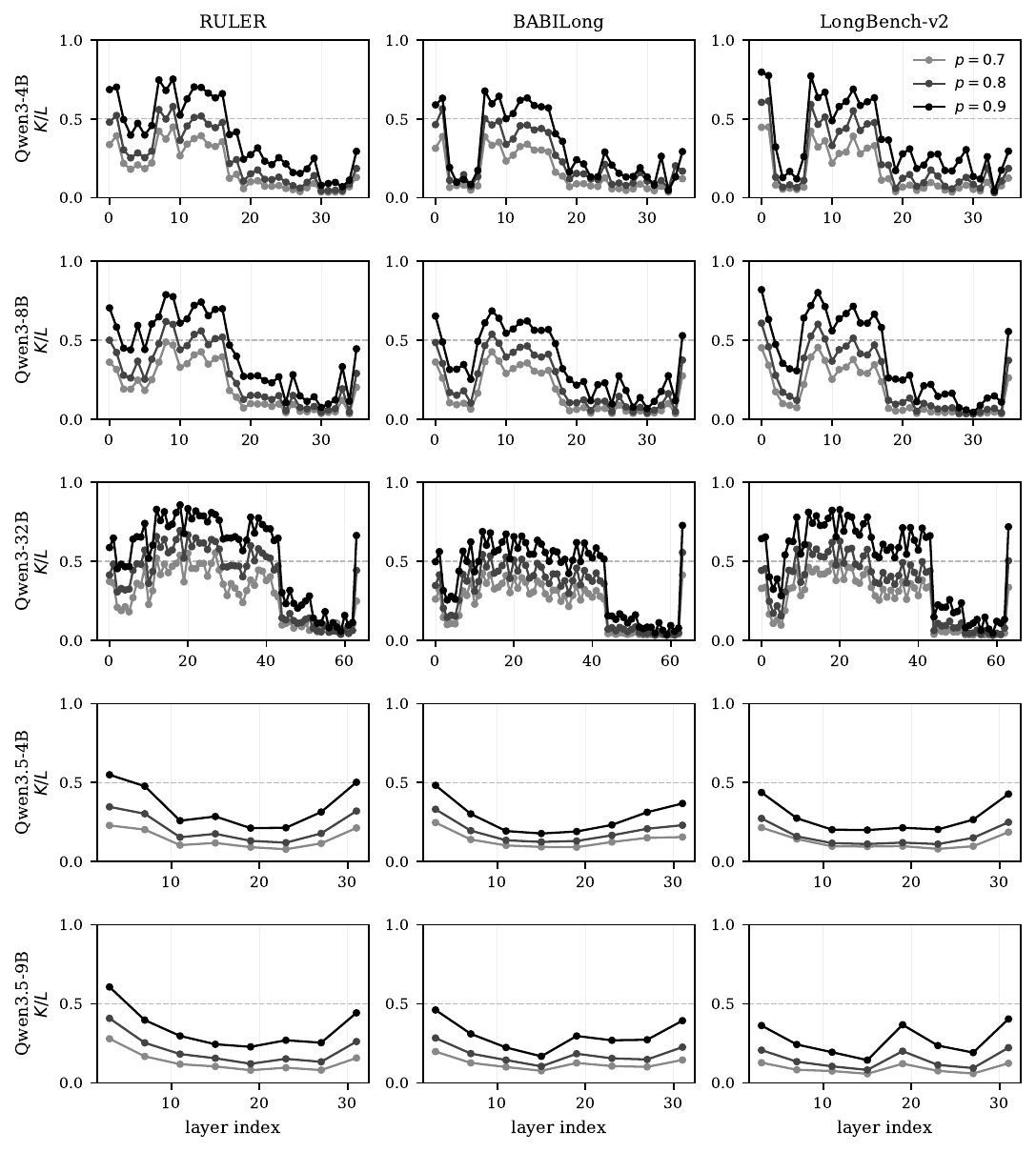}
\caption{\textbf{Per-layer $K/L$ across datasets and top-$p$ values}, 16K context. Rows: backbones; columns: datasets; lines within each panel: $p \in \{0.7, 0.8, 0.9\}$ (light $\to$ dark). Dashed reference at $K/L{=}0.5$.}
\label{fig:perlayer-app}
\end{figure}

Figure~\ref{fig:perlayer-app} extends the per-layer plot across datasets and top-$p$ settings. The signature is dataset-robust: each row has the same shape across RULER, BABILong, and LongBench-v2. It is also budget-robust: the three top-$p$ curves within a panel are nearly parallel, shifted up or down. Both observations support the claim in Section~\ref{sec:analysis-perlayer} that the asymmetry is structural to the backbone rather than a property of the workload.

\section{\Selector{} distribution heatmaps}
\label{app:distribution-shape}

\begin{figure}[t]
\centering
\includegraphics[width=\linewidth]{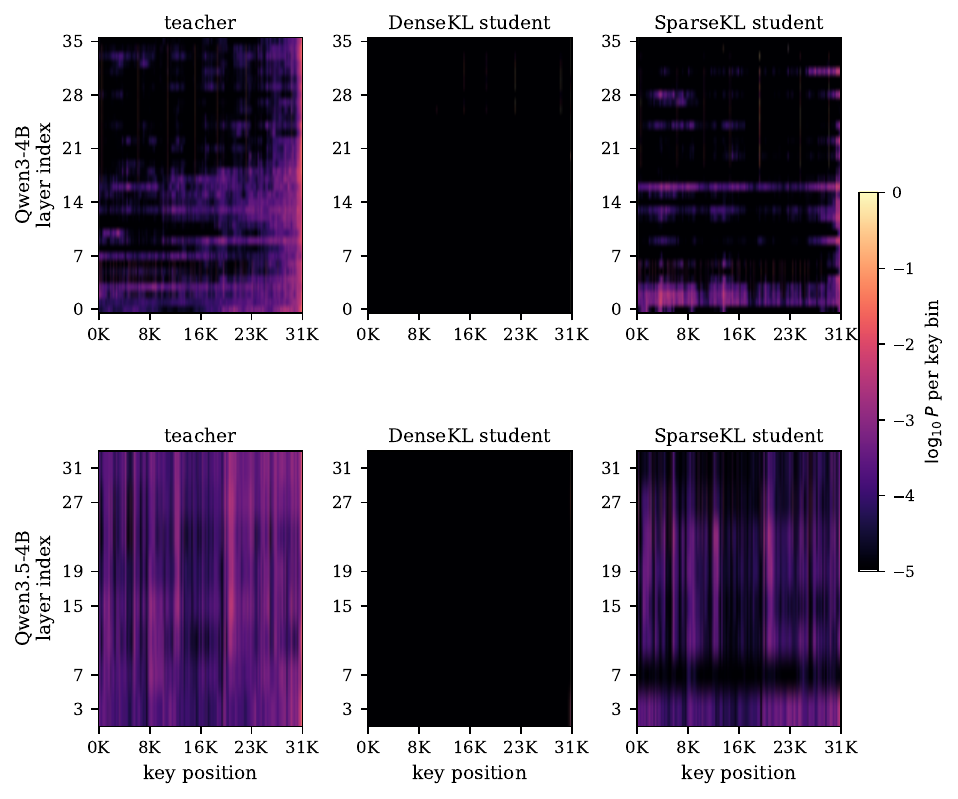}
\caption{\textbf{Last-query softmax distributions, [layer $\times$ key bin], $L = 32$K.} Top row: Qwen3-4B (36 full-attention layers, $K_{\text{train}}{=}8$K). Bottom row: Qwen3.5-4B (8 full-attention layers, $K_{\text{train}}{=}2$K). Columns: teacher / DenseKL student / SparseKL student. Colour: $\log_{10} P$ per bin (lower bound $-5$). DenseKL collapses essentially all mass into a single bin per query (panel reads as black with sparse bright pixels at the right edge), while SparseKL preserves the teacher's diffuse pattern.}
\label{fig:distribution-app}
\end{figure}

Figure~\ref{fig:distribution-app} compares the teacher distribution against the DenseKL and SparseKL students at $L = 32$K on Qwen3-4B and Qwen3.5-4B. The teacher distributions are diffuse: mass spread across content positions, with a recency stripe at the right edge. DenseKL students collapse the mass into a single bin per query (the panels read as black with bright pixels only at the recency edge); SparseKL students recover the teacher's spread.

We capture one 32K FineWeb-Edu forward pass per checkpoint and bin the last-query softmax into 1024 buckets. Qwen3-4B uses production $K_{\text{train}}{=}8$K checkpoints; Qwen3.5-4B uses the matched-$K$ pair at $K_{\text{train}}{=}2$K (no $8$K DenseKL Qwen3.5 was trained).

\begin{figure}[p]
\centering
\includegraphics[width=\linewidth,height=0.85\textheight,keepaspectratio]{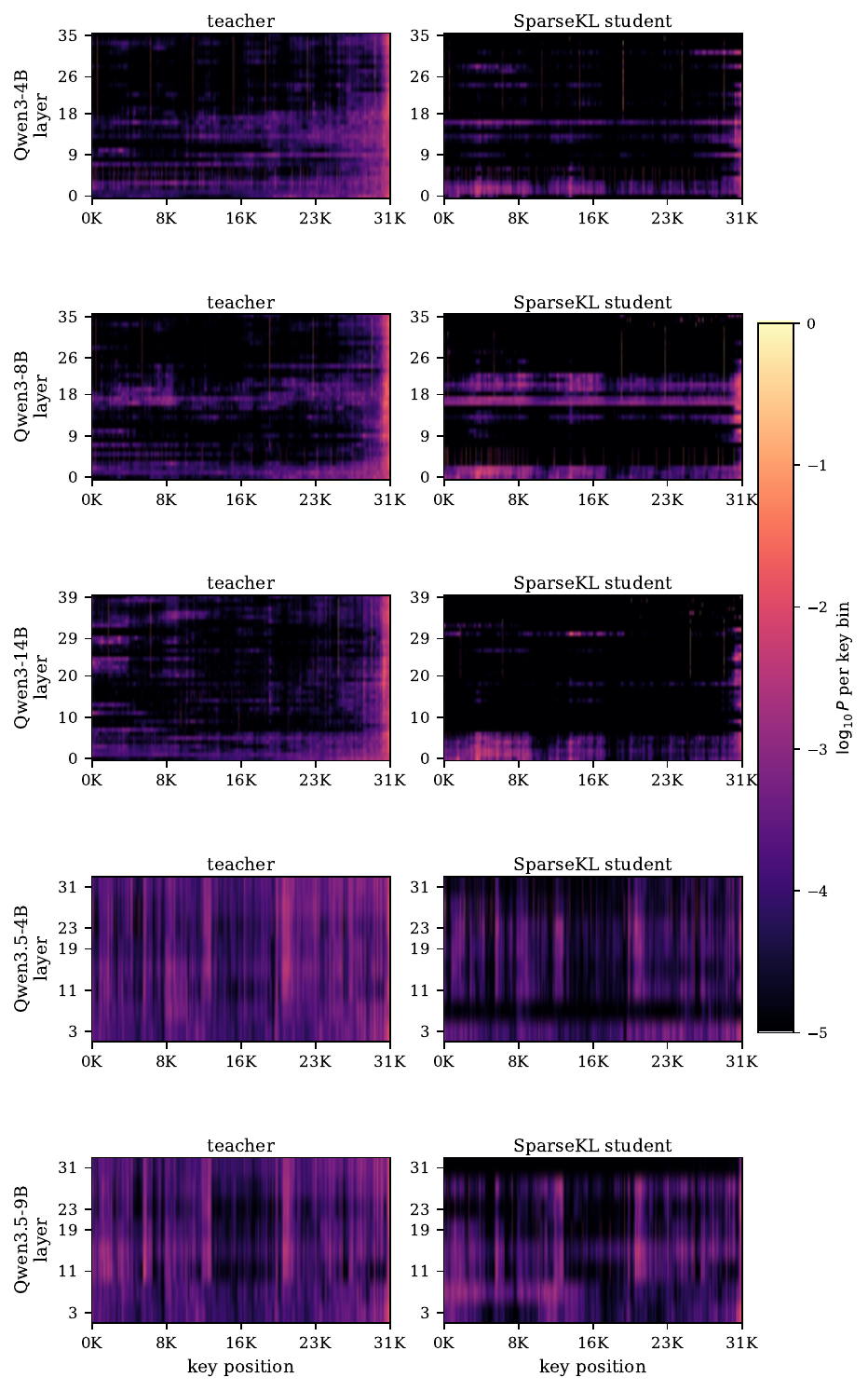}
\caption{\textbf{Teacher vs.\ SparseKL student across the backbone family.} Each row is one backbone; columns are the frozen-backbone dense attention (teacher, left) and the trained SparseKL \selector{} (right). Cells are last-query [layer $\times$ key bin] softmax heatmaps at $L = 32$K, colour $\log_{10} P$. Across Qwen3-\{4B, 8B, 14B\} and Qwen3.5-\{4B, 9B\}, the SparseKL panel recovers the corresponding teacher's spread; the recipe transfers across scale and architecture.}
\label{fig:distribution-models}
\end{figure}

Figure~\ref{fig:distribution-models} extends the comparison across the backbone family. The teacher pattern differs by scale and architecture: Qwen3.5 hybrids carry only eight full-attention layers, while the dense Qwen3 family spans 36--40. Inside each row the SparseKL panel tracks its teacher's spread closely. This is the visual evidence behind the cross-backbone transfer claim in Section~\ref{sec:analysis-distribution}.

\section{\Selector{} quantization}
\label{app:quant}

Table~\ref{tab:quant} reports accuracy at five \selector{} K-cache precisions. The K-cache tolerates 4-bit microscale storage with negligible accuracy loss.

\begin{table}[t]
\centering
\small
\setlength{\tabcolsep}{4pt}
\begin{tabular}{ll cccc}
\toprule
Backbone & Precision & RULER & BABILong & LongBench-v2 & Mean \\
\midrule
\multirow{5}{*}{Qwen3-4B} & BF16 & 0.975 (0.013) & 0.323 (0.028) & 0.270 (0.047) & 0.523 (0.019) \\
 & INT8 & 0.975 (0.013) & 0.317 (0.028) & 0.270 (0.046) & 0.520 (0.018) \\
 & FP8 & 0.975 (0.013) & 0.317 (0.026) & 0.270 (0.047) & 0.520 (0.018) \\
 & INT4 & 0.969 (0.013) & 0.317 (0.026) & 0.270 (0.048) & 0.518 (0.019) \\
 & FP4 & 0.975 (0.013) & 0.320 (0.026) & 0.270 (0.047) & 0.522 (0.018) \\
\midrule
\multirow{5}{*}{Qwen3-8B} & BF16 & 0.981 (0.011) & 0.350 (0.028) & 0.315 (0.051) & 0.549 (0.020) \\
 & INT8 & 0.975 (0.013) & 0.343 (0.027) & 0.315 (0.049) & 0.544 (0.019) \\
 & FP8 & 0.975 (0.013) & 0.340 (0.027) & 0.315 (0.051) & 0.543 (0.020) \\
 & INT4 & 0.981 (0.011) & 0.327 (0.028) & 0.315 (0.049) & 0.541 (0.019) \\
 & FP4 & 0.981 (0.010) & 0.340 (0.026) & 0.315 (0.049) & 0.545 (0.019) \\
\midrule
\multirow{5}{*}{Qwen3-32B} & BF16 & 0.994 (0.006) & 0.363 (0.029) & 0.360 (0.051) & 0.572 (0.020) \\
 & INT8 & 0.994 (0.006) & 0.390 (0.027) & 0.360 (0.050) & 0.581 (0.019) \\
 & FP8 & 0.994 (0.006) & 0.373 (0.028) & 0.360 (0.050) & 0.576 (0.019) \\
 & INT4 & 0.994 (0.006) & 0.397 (0.027) & 0.360 (0.050) & 0.583 (0.019) \\
 & FP4 & 0.988 (0.009) & 0.403 (0.028) & 0.371 (0.053) & 0.587 (0.020) \\
\bottomrule
\end{tabular}
\caption{\textbf{\Selector{} quantization across the Qwen3 family.} Top-$p{=}0.9$ accuracy with the \selector{}'s queries and key cache stored at five precisions, on Qwen3-4B, -8B, and -32B. Cells are mean accuracy (bootstrap standard error in parentheses, Appendix~\ref{app:bootstrap}). Every quantized variant stays within bootstrap error of the BF16 reference on the dataset-mean column at every backbone. \Selector{} K-cache footprint per precision: INT8 / FP8 store at $0.500\times$ BF16; INT4 / FP4 use group-32 microscale at $4.5$ bits per element ($0.281\times$). The backbone KV cache is unchanged; the \selector{} learns under BF16 and is quantized at inference only.}
\label{tab:quant}
\end{table}

\section{Use Of AI Assistants}
AI assistance was used for polishing, copy-editing, and code review during manuscript preparation. All experimental design, scientific claims, numerical results, and final wording were authored and verified by the human authors.

\end{document}